# The Key Concepts of Ethics of Artificial Intelligence

A Keyword based Systematic Mapping Study

Ville Vakkuri

[https://orcid.org/0000-0002-1550-1110]

Faculty of Information Technology

University of Jyväskylä

Jyväskylä, Finland

ville.vakkuri@jyu.fi

Pekka Abrahamsson

[https://orcid.org/0000-0002-4360-2226]

Faculty of Information Technology

University of Jyväskylä

Jyväskylä, Finland

pekka.abrahamsson@jyu.fi

**Abstract** — The growing influence and decision-making capacities of Autonomous systems and Artificial Intelligence in our lives force us to consider the values embedded in these systems. But how ethics should be implemented into these systems? In this study, the solution is seen on philosophical conceptualization as a framework to form practical implementation model for ethics of AI. To take the first steps on conceptualization main concepts used on the field needs to be identified. A keyword based Systematic Mapping Study (SMS) on the keywords used in AI and ethics was conducted to help in identifying, defying and comparing main concepts used in current AI ethics discourse. Out of 1062 papers retrieved SMS discovered 37 re-occurring keywords in 83 academic papers. We suggest that the focus on finding keywords is the first step in guiding and providing direction for future research in the AI ethics field.

## 1. INTRODUCTION

By reviewing the latest accomplishment and increasing implementation of Autonomous systems (AS) and Artificial Intelligence (AI) systems have become more influential in our lives. By growing influence ethical questions related to these systems have become more and more obvious and actual.

For example, looking biased algorithms in social media[1], decision making systems of autonomous cars[2], or even social effects of automatization in whole transportation ecosystems like autonomous maritime[3] it is clear that system development is not anymore only about technological or engineering question. AI and AS are already in the surrounding world among us and the need of implementing ethics and our values into these systems is urgent.

Concerning ethics as a part of system design has also gained attention from governmental and standardization level, such as Federal Ministry of Transport and Digital Infrastructure in Germany[4] and IEEE[5]. The academic discussion on the relation of AI and ethics has been ongoing for decades, but the development of systems and ethical research have only slightly crossed[6]. The ethical research has been mainly focused on the potential of AI on theoretical level [7]. So, the question still remains open on application level: How ethics should be implemented in practice into these systems?

There can be little ethical implementation without understanding the consequences of developers' own actions, open dialogue and ethical aspects considered in AI and autonomous system development, because of the multidisciplinary nature of AI ethics development [8]. As a solution for understanding the field of ethics of AI, philosophical conceptualization should be used. This method allows to discuss and to form cross-disciplinary definitions for key concepts and also initiate productive dialog merging philosophical and technological views to produce a common framework for implementing ethics in AI.

The goal of this paper is to identify and categorize keywords used in academic papers in the current AI ethics discourse and by that take first steps to identify, define and compare main concepts and terms used in discourse. To find the relevant papers and keywords, a preliminary Systematic Mapping Study was conducted with the following focus:

- o  Recognize keywords used in the field
- o  Extract potential keywords for future research
- o  Compare keywords to proposed concepts in academic literature

The Systematic Mapping Study based on keywords reveals 37 re-occurring author keywords found in 83 academic papers that are found from an initial set of 1062 papers in the field of AI and ethics. Cause of the preliminary nature of this study as Systematic Mapping Study it does not provide full comprehensive picture of the primary studies in the area, but it provides an important standpoint and relevant tools for future research on AI and ethics. By understanding the used concepts, research can shift from discussing concepts to defining concepts and this way aids the need of practical implementation of ethics into AI systems.

This paper is organized as follows: Section 2 describes the background and related work; Section 3 describes the research methodology and conducted keyword based search; Section 4 findings; Section 5 concludes the paper by discussing the findings with general presentation of AI ethics and summarizes the answers for research questions to set guidelines for future work.

## 2. BACKGROUND

The ethical discussion of Artificial Intelligence has been present from the start for AI research, but instead of focusing on the real use cases, the focus has been mainly on the theoretical work discussing the possibilities and future impacts of AI. In recent years there has been a major change in discussion of AI related ethics when new level of capabilities of AI have become reality and more influential in our lives due to the resent breakthroughs in AI development. Availability of low cost computing power and innovation like Big Data technologies have made AI more useable in solving complicated problems. [9] One milestone of AI development can be seen in year 2012 when Google's large-scale deep learning experimentation on brain simulation using 16000 CPU cores and deep learning was conducted[10]. The experiment significantly improved the state of the art on a standard image classification test. This year also serves as the starting point for the current AI ethics discussion in the context of this study.

Even though the academic discussion on the relation of AI and ethics has been ongoing for decades, there is no commonly shared definition of what AI ethics is or even how it should be named. As the defining concept, Machine ethics has arisen out the discussion but it has also been criticized. There has been a heated discussion on how does the concept of machine ethics also cover and include new branches in AI related ethics. [7, 11, 12]

There is only a handful of books that have comprehensive presentations covering the ethical issues of AI, such as Towards a Code of Ethics for Artificial Intelligence that mainly focuses on professional ethics[9]; robot ethics 2.0 covers ethics related to embodied AI[13] and Machine Ethics prior to the current discussion[14]. For defining the field of AI related ethics so called "six hot topics" have proposed [15]. The problem with these categorically wide topics is that they are not necessarily comprehensive or clear enough and not in balance with the overall discussion. Importantly, they are also not necessarily scientifically founded. For example of the wide scope of AI ethics discourse, the first AAAI/ACM Conference on AI, Ethics, and Society held in 2018 had broad set of 12 different topics from technical to social sciences[16].

Besides defining relevant concepts for a crucial problem in practical implementation of AI ethics is the limited co-operation and communication between the developers of the AI systems and ethics researchers.[6, 11] To reach the practical implementation of AI ethics, a multidisciplinary research approach is needed where AI developers can also see the use of ethics and results of the philosophical research on a practical level.

## 3. RESEARCH METODOLOGY AND MAPPING

As a multidisciplinary research area AI ethics covers a wide range of topics and the discussion of definitions still endures. To gain a better understanding of the research area, a Systematic Mapping Study was chosen as a research method due to its capability to deal with wide and loosely defined areas of study. SMS aims at producing an overview of the field and reveals concretely which topics have been covered to a certain extent. The present study is a keyword based systematic mapping study. Two main guidelines for systematic mapping study were combined aiming at recognizing primary studies and the used keywords therein. We consider this study, however, to be the first step since the mapping process is not executed to its full length. We needed first to gain a better understanding of relevant keywords for the PICO (Population, Intervention, Comparison and Outcomes) process. [17, 18]

### A. Definition of Research Questions

The main research question for the present study is: What are the main author keywords used in academic papers in the current AI ethics discussion. To answer this question, four sub-questions were formed:

- o Q1 What are the author keywords used?
- o Q2 Which of the keywords are re-occurring and in which pattern?
- o Q3 How can the author keywords be classified?
- o Q4 How do the used keywords reflect the proposed concepts in academic AI ethics literature?

The purpose of Q1 is to produce a preliminary picture of the keywords used in the identified papers and gathering information together. Q2 aims at recognizing the main keywords by means of a quantitative analysis of the variance and appearance in the identified papers whereas Q3 aims at providing qualitative classification of the used keywords. With Q4 the intention is to understand how keywords fit into proposed concepts, how comprehensive they are and what type of new concepts they can potentially offer.

*B. Conducted Search*

Keywords were identified by conducting keyword search in selected scientific databases. The search string was formed from the main research question by combining both key concepts artificial intelligence/AI + ethics. The suggested PICO process was not used to identify search string keywords because of the lack of shared concepts in AI ethics for the reasons argued earlier.

The selected scientific databases on which search was performed are shown in Table II, along with the number of publications retrieved from each database (in the 11th of March, 2018). The selection of databases were guided by the need to gain a wide coverage of the multidisciplinary nature of AI research and databases ability to handle advanced queries. The used set of keyword search strings were customized as shown in Table I to adapt to the syntax of the particular database. Web of Science and ProQuest databases do not have specified search term for author keyword, therefore keyword including the topic and subject fields were used in the search queries.

TABLE I. DATABASES AND RESEACH STRINGS

| Database | Search String |
|---|---|
| IEEE Xplore (ieeexplore.ieee.org) | (("Author Keywords":ethics) AND "Author Keywords":artificial intelligence) |
| ACM Digital Library (dl.acm.org/advsearch.cfm) | keywords.author.keyword:( +"artificial intelligence" +ethics) |
| Scopus (www.scopus.com) | KEY ( "artificial intelligence" OR ai AND ethics ) |
| Web of Science (wokinfo.com) | TOPIC: ("artificial intelligence") AND TOPIC: (ethics) |
| ProQuest (www.proquest.com) | (SU.exact("ETHICS") AND SU.exact("ARTIFICIAL INTELLIGENCE")) |

TABLE II. DATABASES AND RETRIEVED PAPERS

| Database | Papers |
|---|---|
| IEEE Xplore (ieeexplore.ieee.org) | 15 |
| ACM Digital Library (dl.acm.org/advsearch.cfm) | 27 |
| Scopus (www.scopus.com) | 320 |
| Web of Science (wokinfo.com) | 83 |
| ProQuest (www.proquest.com) | 617 |
| **Total retrieved** | **1062** |

*C. Screening of Relevant Papers*

Papers were included from search results by following criteria:

- o Scholarly Journal articles
- o Written in English language
- o Part of current discussion, published 2012 or after
- o Related to ethics and artificial intelligence or related technologies
- o Full-text available for reviewing
- o Author keywords available for extraction

Pre-exclusion of document type, source type and article language was done automatically in databases, see Table III. From databases five different results lists were exported and combined to reference management tool RefWorks resulting list of 588 papers. For duplicate exclusion each papers metadata and title were reviewed with aid of the reference management tool. In manual metadata analysis, papers published before 2012 were excluded. In addition, non-scholarly journal articles, for example popular articles, which were not detected in pre-exclusion phase, were excluded in the manual screening process. In in-depth review of the remaining papers, abstracts were analyzed to determinate whether the paper is related to ethics and artificial intelligence or related technologies. In the last iteration of exclusion, papers were excluded if full-text and author keywording were not available. Resulting 83 papers included. Screening process and steps can be seen in Table III and distribution by year in Fig. 1.

TABLE III. EXCLUDED PAPERS

| Rationale | Amount |
|---|---|
| **Pre exclusion in Database:** | |
| Document type | -365 |
| Source type | -96 |
| Not in English language | -13 |
| **Manual exclusion** | -148 |
| Duplicate | -237 |
| Published before 2012Academic settings or Document type | -76 |
| Not in English language | -1 |
| No Full-text available | -27 |
| No author keywording available | -16 |
| **Total retrieved** | **1062** |
| **Total excluded** | **979** |
| **Total included** | **83** |

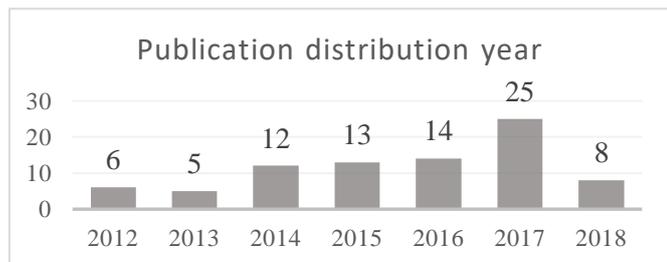

Fig. 1. Included publications distribution by year.

## 4. FINDINGS

For this study, the listing of the included keywords worked as a data extraction and no further keywording was conducted. To answers research questions Q1 and Q2, the keywords were listed and counted resulting in total of 324 different keywords in 83 papers. 37 of the 324 used keywords were re-occurring in two or more papers. Most frequently used keywords were Artificial intelligence/AI and ethics. This is a natural result due to the terms used in research strings, and therefore does not provide new information. These keywords were excluded from the listing. Re-occurring keywords and papers where these keywords were used can be seen on Table IV. The usage of keywords has considerable variance in incidence and spelling such as "Roboethics" and "Robot ethics" that may hinder search result. The variance in used keywords for one topic such as "Autonomous vehicle", "Driverless cars", "Self-driving cars" can be seen also as example of the immaturity of shared terms and undisclosed discussion what terms should be used in specific context.

TABLE IV. RE-OCCURRING KEYWORDS

| Keyword | n | Found in |
| --- | --- | --- |
| Machine ethics | 16 | [7, 12, 19-32] |
| Robotics | 11 | [20, 27, 33-41] |
| Robots | 7 | [37, 39, 42-46] |
| Autonomy | 5 | [23, 30, 44, 47, 48] |
| Responsibility | 5 | [22, 36, 49-51] |
| Roboethics | 5 | [22, 35, 41, 52, 53] |
| Robot ethics | 4 | [32, 54-56] |
| Artificial agents | 4 | [30, 50, 52, 57] |
| artificial general intelligence | 3 | [7, 24, 29] |
| Artificial moral agents | 3 | [32, 54, 58] |
| Automation | 3 | [34, 46, 59] |
| Consciousness | 3 | [31, 60, 61] |
| existential risk | 3 | [29, 62, 63] |
| free will | 3 | [52, 61, 64] |
| Moral agency | 3 | [53, 65, 66] |
| Moral patiency | 3 | [53, 55, 65] |
| Self-driving cars | 3 | [58, 59, 67] |
| Value alignment | 3 | [19, 68, 69] |
| AI ethics | 3 | [19, 70, 71] |
| Anthropocentrism | 2 | [31, 36] |
| Artificial morality | 2 | [22, 58] |
| Autonomous agents | 2 | [22, 71] |
| Autonomous vehicle | 2 | [72, 73] |
| Driverless cars | 2 | [74, 75] |
| friendly AI | 2 | [29, 68] |
| Human rights | 2 | [20, 36] |
| human-robot interaction | 2 | [54, 56] |
| Information technology | 2 | [76, 77] |
| Machine Intelligence | 2 | [19, 76] |
| Moral status | 2 | [31, 70] |
| Personhood | 2 | [50, 78] |
| Regulation | 2 | [66, 79] |
| Rights | 2 | [49, 50] |
| Self | 2 | [47, 61] |
| Superintelligence | 2 | [24, 63] |
| Trust | 2 | [30, 57] |
| Virtue ethics | 2 | [46, 52] |

The 37 re-occurring keywords where classified into 9 categories as shown in Table V. Classification of keywords was formed following four step process: 1) Linguistic similarity of keywords, for example similarity in spelling. 2) Ontological similarity of keyword as assumed reference for same concept. 3) Family resemblance of keywords. 4) Similarity in usage, from abstract to specific. After classification describing names were given to formed categories. [80]

The idea of classification was to outline re-occurring topics from the vast variance of keywords. This classification produced a comparative set of more general topics relevant to AI ethics. By looking at the keywords listed it is surprising that different branches of AI such as Machine learning, Natural Language Processing or Pattern recognition were not found in the set keywords. This may imply that the relevant ethical discussion is done under separate AI branches and cannot be found through AI/artificial intelligence keyword.

Academic literature shows similarities in recurring terms when comparing keywords and the formed categories to proposed concepts and topics, but keyword listing is in some parts also partial. For example, technology based keywords and topics are underrepresented such as bias issues, fairness, transparency and controlling AI. Also socioethical topics like impact on society or workforce are lacking. [9, 13] Comparison of keyword classification reveals topics that are quite commonly shared in literary. Found keywords can be classified under the known topics even specified formulation of keywords in some parts varies considerably.

TABLE V. FORMED CATEGORIES

| Category | Keywords |
|---|---|
| Conceptual | AI ethics, Machine ethics, Information technology Sports ethics, Virtue ethics, Friendly AI |
| Robotics | Robotics, Robots, Roboethics, Robot ethics, Automation |
| Generally Philosophical And Ethical | Autonomy, Autonomous agents, free will, Moral agency, Moral patiency, Moral status, Trust, Anthropocentrism Personhood, Self |
| AI specified Philosophical And Ethical | Artificial agents, Artificial moral agents Artificial morality |
| Law and Regulation | Regulation, Rights, Responsibility, Human rights |
| Autonomous vehicle | Autonomous vehicle, Driverless cars, Self-driving cars |
| AGI and AI risk | artificial general intelligence, superintelligence, existential risk |
| Human cognition | Intelligence, Consciousness, Machine Intelligence, human-robot interaction |
| Technology based | Value alignment |

## 5. DISCUSSION & CONCLUSION

This study provided a set of AI ethics related keywords and listing of 37 re-occurring author keywords found in 83 academic papers. Re-occurring keywords where classified into 9 categories based on conceptual similarities of keywords to more general topics relevant to AI ethics. Keywords and formed categories where compared to concepts provided in academic literature to evaluate coverage of the systematic mapping study and listing. Three main differences were discovered: Lack of different branches of AI in keywords, technology based keywords have only minor role and there is a great variance in formulation of keywords even though keywords can be classified under the known topics. Recommendation for future research and systematic mapping studies: Different AI branches and different formulation for keywords extracted from known topics should be included in the keyword extraction process.

Keyword based systematic mapping study method used in this study has several weaknesses. Due to the focus on the keywords only, no primary studies of the field of AI ethics where recognized. The relevance of papers was evaluated in exclusion process and in the prevalence of keywords in the papers. Neither definitions of concepts that keywords represented where not analyzed. Despite the weaknesses, keyword based approach allowed to cover wide and loosely defined field of AI ethics to produce understanding of relevant keywords where no prior listing was available. This preliminary work also helps future systematic mapping studies by providing relevant keywords on AI ethics.

With wide variety of papers and keywords from different areas concerning AI ethics this study revealed that defining the field of AI ethics is still a challenging task. The comprehensive presentations have done a valuable work on setting definitions for expanding field of AI ethics. There is still a substantial amount of work to be done in the area. These presentations are not all inclusive and more comprehensive works are needed on the topic discussed on this paper. For example, by looking at the occurrence of different keywords, papers have different stress in different topics than comprehensive presentations have. Overall there is still research needed in the field of AI ethics on the concepts as such to see where AI ethics discourse is developing and how concepts can aid the need of practical implementation of ethics into AI systems.

## 6. REFERENCES

[1] W. Knight, "Biased Algorithms Are Everywhere, and No One Seems to Care," 2017, https://www.technologyreview.com/s/608248/biased-algorithms-are-everywhere-and-no-one-seems-to-care/, Retrieved April 21, 2018.

[2] J.D. Greene, "Our driverless dilemma," Science, vol. 352, pp. 1514-1515, 2016.

[3] Anonymous, "About Autonomous Shipping," https://www.oneseaecosystem.net/about/about-autonomous-shipping/, Retrieved April 21, 2018.

[4] BMVI, "Ethics Commission's complete report on automated and connected driving," 2017, https://www.bmvi.de/goto?id=354980, Retrieved April 21, 2018.

[5] IEEE Global Initiative, "The IEEE Global Initiative on Ethics of Autonomous and Intelligent Systems," http://standards.ieee.org/develop/indconn/ec/autonomous_systems.html, Retrieved April 21, 2018.

[6] C. Allen, W. Wallach and I. Smit, "Why Machine Ethics?" IEEE Intelligent Systems, vol. 21, pp. 12-17, 2006.

[7] M. Brundage, "Limitations and risks of machine ethics," Journal of Experimental & Theoretical Artificial Intelligence, vol. 26, pp. 355-372, 2014.

[8] C. Mayer, "Developing Autonomous Systems in an Ethical Manner," Issues for Defence Policymakers, vol. 65, 2015.

[9] P. Boddington, "Towards a Code of Ethics for Artificial Intelligence," Cham: Springer, 2017.

[10] J, Dean, "Using large-scale brain simulations for machine learning and A.I." vol. 2018, June 26, 2012.

[11] V. Charisi, L. Dennis, M.F.R. Lieck, A. Matthias, M.S.J. Sombetzki, A.F. Winfield and R. Yampolskiy, "Towards moral autonomous systems," arXiv Preprint arXiv:1703.04741, 2017.

[12] R. Yampolskiy, "Safety Engineering for Artificial General Intelligence," Topoi, vol. 32, pp. 217-226, 2013.

[13] P. Lin, "Robot Ethics 2.0: From Autonomous Cars to Artificial Intelligence," Oxford: Oxford University Press, 2017.

[14] M. Anderson, S. L. Anderson, "Machine ethics," New York: Cambridge University Press, 2011.

[15] D. Zeng, "AI Ethics: Science Fiction Meets Technological Reality," Intelligent Systems, IEEE, vol. 30, pp. 2-5, 2015.

[16] Conference on Artificial Intelligence, Ethics and Society, "Call for papers (Main track)," http://www.aies-conference.com/call-for-papers/, Retrieved April 21, 2018.

[17] B. Kitchenham and S. Charters, "Guidelines for performing Systematic Literature Reviews in Software Engineering," Keele University and Durham University Joint Report, 2007.

[18] K. Petersen, "Guidelines for conducting systematic mapping studies in software engineering: An update," Information and Software Technology, vol. 64, pp. 1-18, 2015.

[19] K. Bogosian, "Implementation of Moral Uncertainty in Intelligent Machines," Minds and Machines, vol. 27, pp. 591-608, 2017.

[20] H. Ashrafian, "AIonAI: A Humanitarian Law of Artificial Intelligence and Robotics," Sci.Eng.Ethics, vol. 21, pp. 29-40, 2015.

[21] J. Cervantes, "Autonomous Agents and Ethical Decision-Making," Cognitive Computation, vol. 8, pp. 278-296, 2016.

[22] G. Dodig Crnkovic, "Robots: ethical by design," Ethics and Information Technology, vol. 14, pp. 61-71, 2012.

[23] A. Etzioni, "The ethics of robotic caregivers," Interaction Studies, vol. 18, pp. 174-190, 2017.

[24] B. Goertzel, "GOLEM: towards an AGI meta-architecture enabling both goal preservation and radical self-improvement," Journal of Experimental & Theoretical Artificial Intelligence, vol. 26, pp. 391-403, 2014.

[25] M. Graves, "Shared Moral and Spiritual Development Among Human Persons and Artificially Intelligent Agents," Theology and Science, vol. 15, pp. 333-351, 2017.

[26] D. Gunkel, "A Vindication of the Rights of Machines," Philosophy & Technology, vol. 27, pp. 113-132, 2014.

[27] T. Hauer, "Society and the Second Age of Machines: Algorithms Versus Ethics," Society, pp. 1-7, 2018.

[28] R.M. Omari and M. Mohammadian, "Rule based fuzzy cognitive maps and natural language processing in machine ethics," Journal of Information, Communication and Ethics in Society, vol. 14, pp. 231-253, 2016.

[29] K. Sotala, "Responses to catastrophic agi risk: a survey," Phys.Scripta, vol. 90, pp. 018001, 2014.

[30] H.T. Tavani, "Levels of Trust in the Context of Machine Ethics," Philosophy & Technology vol. 28.1, pp. 75-90, 2015.

[31] S. Torrance, "Artificial agents and the expanding ethical circle," AI & Society, vol. 28, pp. 399-414, 2013.

[32] A. van Wynsberghe and S. Robbins, "Critiquing the Reasons for Making Artificial Moral Agents," Sci.Eng.Ethics, 2018.

[33] J. Bryson, "Standardizing Ethical Design for Artificial Intelligence and Autonomous Systems," Computer, vol. 50, pp. 116-119, 2017.

[34] M. Coeckelbergh, "The tragedy of the master: automation, vulnerability, and distance," Ethics and Information Technology, vol. 17, pp. 219-229, 2015.

[35] G. Ghilardi, "Post-human and scientific research: how engineering carried out the project," Cuadernos De Bioetica : Revista Oficial De La Asociacion Espanola De Bioetica Y Etica Medica, vol. 25, pp. 379, 2014.

[36] L. Hin-Yan, "From responsible robotics towards a human rights regime oriented to the challenges of robotics and artificial intelligence," Ethics and Information Technology, pp. 1-13, 2017.

[37] S. Russell, "Robotics: Ethics of artificial intelligence," Nature, vol. 521, pp. 415, 2015.

[38] B. Schafer, "A fourth law of robotics? Copyright and the law and ethics of machine co-production," Artificial Intelligence and Law, vol. 23, pp. 217-240, 2015.

[39] M. Szollosy, "Rapporteur's report," Connect.Sci., vol. 29, pp. 254-263, 2017.

[40] G. Tamburrini, "On the ethical framing of research programs in robotics," Ai & Society, vol. 31, pp. 463-471, 2016.

[41] A. Theodorou, "Designing and implementing transparency for real time inspection of autonomous robots," Connect.Sci., vol. 29, pp. 230-241, 2017.

[42] J.J. Bryson, "Of, for, and by the people: the legal lacuna of synthetic persons.(Special Issue: Machine Law)," Artificial Intelligence and Law, vol. 25, pp. 273, 2017.

[43] F. Javier Lopez Frias, "Will robots ever play sports?" Sport, Ethics and Philosophy, vol. 10, pp. 67-82, 2016.
[44] D. Johnson, "Reframing AI Discourse," Minds and Machines, vol. 27, pp. 575-590, 2017.
[45] P. Kopacek, "Roboethics," IFAC Proceedings Volumes, vol. 45, pp. 67-72, 2012.
[46] S. Vallor, "Moral Deskilling and Upskilling in a New Machine Age: Reflections on the Ambiguous Future of Character," Philosophy & Technology, vol. 28.1, pp. 107-12, 2015.
[47] M. Coeckelbergh, "Pervasion of what? Techno--human ecologies and their ubiquitous spirits.(Report)," AI & Society, vol. 28, pp. 55, 2013.
[48] E. Colombetti, "Contemporary post-humanism: technological and human singularity," Cuadernos De Bioetica : Revista Oficial De La Asociacion Espanola De Bioetica Y Etica Medica, vol. 25, pp. 367, 2014.
[49] H. Ashrafian, "Artificial Intelligence and Robot Responsibilities: Innovating Beyond Rights," Sci.Eng.Ethics, vol. 21, pp. 317-326, 2015.
[50] M. Laukyte, "Artificial agents among us: Should we recognize them as agents proper?," Ethics and Information Technology, vol. 19, pp. 1-17, 2017.
[51] H. Liu, "Irresponsibilities, inequalities and injustice for autonomous vehicles," Ethics and Information Technology, vol. 19, pp. 193-207, 2017.
[52] V. Galanos, "Singularitarianism and schizophrenia," Ai & Society, vol. 32, pp. 573-590, 2017.
[53] D. Gunkel, "Introduction to the Special Issue on Machine Morality: The Machine as Moral Agent and Patient," Philosophy & Technology, vol. 27, pp. 5-8, 2014.
[54] T. Arnold, "Against the moral Turing test: accountable design and the moral reasoning of autonomous systems," Ethics and Information Technology, vol. 18, pp. 103-115, 2016.
[55] E. Neely, "Machines and the Moral Community," Philosophy & Technology, vol. 27, pp. 97-111, 2014.
[56] K. Stowers, "Life or Death by Robot?" Ergonomics in Design: The Quarterly of Human Factors Applications, vol. 24, pp. 17-22, 2016.
[57] F.S. Grodzinsky, "Developing Automated Deceptions and the Impact on Trust," Philosophy & Technology vol. 28.1, pp. 91-105, 2015.
[58] C. Misselhorn, "Artificial Morality. Concepts, Issues and Challenges," Society, pp. 1-9, 2018.
[59] J. Borenstein, "Self-Driving Cars and Engineering Ethics: The Need for a System Level Analysis," Sci.Eng.Ethics, pp. 1-16, 2017.
[60] J.J. Bryson, "A ROLE FOR CONSCIOUSNESS IN ACTION SELECTION," International Journal of Machine Consciousness, vol. 4, pp. 471-482, 2012.
[61] M.R. Waser, "SAFE/MORAL AUTOPOIESIS AND CONSCIOUSNESS," International Journal of Machine Consciousness, vol. 5, pp. 59-74, 2013.
[62] K.S. Gill, "Data Driven Wave of Certainty- a question of ethical sustainability," IFAC PapersOnLine, vol. 49, pp. 117-122, 2016.
[63] K. Sotala, "Superintelligence as a Cause or Cure for Risks of Astronomical Suffering," Informatica, vol. 41, pp. 389, 2017.
[64] L. Frank, "Robot sex and consent: Is consent to sex between a robot and a human conceivable, possible, and desirable?" Artificial Intelligence and Law, vol. 25, pp. 305, 2017.
[65] J. Bryson, "Patiency is not a virtue: the design of intelligent systems and systems of ethics," Ethics and Information Technology, vol. 20, pp. 15-26, 2018.
[66] D. Martin, "Who Should Decide How Machines Make Morally Laden Decisions?" Sci.Eng.Ethics, vol. 23, pp. 951-967, 2017.
[67] A. Etzioni, "Incorporating Ethics into Artificial Intelligence," The Journal of Ethics, vol. 21, pp. 403-418, 2017.
[68] G.P. Sarma, "Mammalian Value Systems," Informatica, vol. 41, pp. 441, 2017.
[69] P. Vamplew, "Human-aligned artificial intelligence is a multiobjective problem," Ethics and Information Technology, vol. 20, pp. 27-40, 2018.
[70] J. Basl, "Machines as Moral Patients We Shouldn't Care About (Yet): The Interests and Welfare of Current Machines," Philosophy & Technology, vol. 27, pp. 79-96, 2014.
[71] M. Wellman, "Ethical Issues for Autonomous Trading Agents," Minds and Machines, vol. 27, pp. 609-624, 2017.
[72] G. Contissa, "The Ethical Knob: ethically-customisable automated vehicles and the law," Artificial Intelligence and Law, vol. 25, pp. 365-378, 2017.
[73] K. Kinjo, "Optimal program for autonomous driving under Bentham- and Nash-type social welfare functions," Procedia Computer Science, vol. 112, pp. 61-70, 2017.
[74] A. Etzioni, "AI assisted ethics," Ethics and Information Technology, vol. 18, pp. 149-156, 2016.
[75] D. Purves, "Autonomous Machines, Moral Judgment, and Acting for the Right Reasons," Ethical Theory and Moral Practice, vol. 18, pp. 851-872, 2015.
[76] M. Loi, "Technological unemployment and human disenhancement," Ethics and Information Technology, vol. 17, pp. 201-210, 2015.
[77] M. Rader, "The jobs of others: "speculative interdisciplinarity" as a pitfall for impact analysis," Journal of Information, Communication and Ethics in Society, vol. 10, pp. 4-18, 2012.
[78] D.R. Lawrence, "Artificial Intelligence," Quarterly of Healthcare Ethics vol. 25.2, pp.250-261, 2016.
[79] I. Rahwan, "Society-in-the-loop: programming the algorithmic social contract," Ethics and Information Technology, vol. 20, pp. 5-14, 2018.
[80] A. Thomasson, "Categories," Mar 7,. 2018, https://plato.stanford.edu/archives/spr2018/entries/categories/, Retrieved April 21, 2018.